\pdfoutput=1

\documentclass[11pt]{article}
\usepackage[dvipsnames, svgnames, x11names]{xcolor}
\usepackage{EMNLP2022}

\usepackage{times}
\usepackage{latexsym}

\usepackage[T1]{fontenc}

\usepackage[utf8]{inputenc}

\usepackage{microtype}

\usepackage{inconsolata}

\usepackage{booktabs}
\usepackage{tabularx}
\usepackage{multirow}
\usepackage{amsmath}
\usepackage{amsfonts}
\usepackage{pgfplots}
\usepackage{pgfplotstable}
\usepackage{subcaption}
\usepackage{pifont}

\usepackage{colortbl}
\usepackage{algorithmic}
\usepackage[ruled,vlined]{algorithm2e}
\usepackage{CJK}

\title{BBTv2: Towards a Gradient-Free Future with Large Language Models}

\author{
Tianxiang Sun\textsuperscript{$\diamondsuit\heartsuit$}\quad \quad
Zhengfu He\textsuperscript{$\diamondsuit$}\quad \quad
Hong Qian\textsuperscript{$\spadesuit$}\\
\bf{
Yunhua Zhou\textsuperscript{$\diamondsuit\heartsuit$}\quad \quad
Xuanjing Huang\textsuperscript{$\diamondsuit\heartsuit$}\quad \quad
Xipeng Qiu\textsuperscript{$\diamondsuit\heartsuit$}\thanks{\ \ \ Corresponding author.}
}\\
\textsuperscript{$\diamondsuit$}School of Computer Science, Fudan University\\
\textsuperscript{$\heartsuit$}Shanghai Key Laboratory of Intelligent Information Processing, Fudan University\\
\textsuperscript{$\spadesuit$}School of Computer Science and Technology, East China Normal University\\
\texttt{\{txsun19,zfhe19,zhouyh20,xjhuang,xpqiu\}@fudan.edu.cn}\quad 
\texttt{hqian@cs.ecnu.edu.cn}
  }

\begin{document}
\maketitle
\begin{abstract}
Most downstream adaptation methods tune all or part of the parameters of pre-trained models (PTMs) through gradient descent, where the tuning cost increases linearly with the growth of the model size.
By contrast, gradient-free methods only require the forward computation of the PTM to tune the prompt, retaining the benefits of efficient tuning and deployment.
Though, past work on gradient-free tuning often introduces gradient descent to seek a good initialization of prompt and lacks versatility across tasks and PTMs.
In this paper, we present BBTv2, an improved version of Black-Box Tuning~\cite{Sun2022BBT}, to drive PTMs for few-shot learning.
We prepend continuous prompts to every layer of the PTM and propose a divide-and-conquer gradient-free algorithm to optimize the prompts at different layers alternately.
Extensive experiments across various tasks and PTMs show that BBTv2 can achieve comparable performance to full model tuning and state-of-the-art parameter-efficient methods (e.g., Adapter, LoRA, BitFit, etc.) under few-shot settings while maintaining much fewer tunable parameters.

\end{abstract}

\section{Introduction}
The past few years have witnessed remarkable progress of large language models (LLMs)~\cite{Devlin2019BERT,Raffel2020T5,Brown2020GPT3}. It has been repeatedly demonstrated that scaling up the model size is promising to achieve better performance. However, the growing model size also leads to a linear increase in tuning cost. Fine-tuning and deploying a separate copy of the LLM for each downstream task become prohibitively expensive. To that end, much effort has been devoted to \textit{parameter-efficient tuning} (PET)~\cite{He2021Unified,Ding2022Delta}, which only tunes a small portion of parameters while keeping most of the parameters of the LLM unchanged. By PET, LLMs can be specialized to a downstream task at inference time by activating a small number of task-specific parameters. Though it is deployment-efficient, tuning the small portion of parameters still requires back-propagation through the entire LLM, which is expensive or even infeasible for many practitioners.

\begin{figure}[t]
    \centering
    \includegraphics[width=\linewidth]{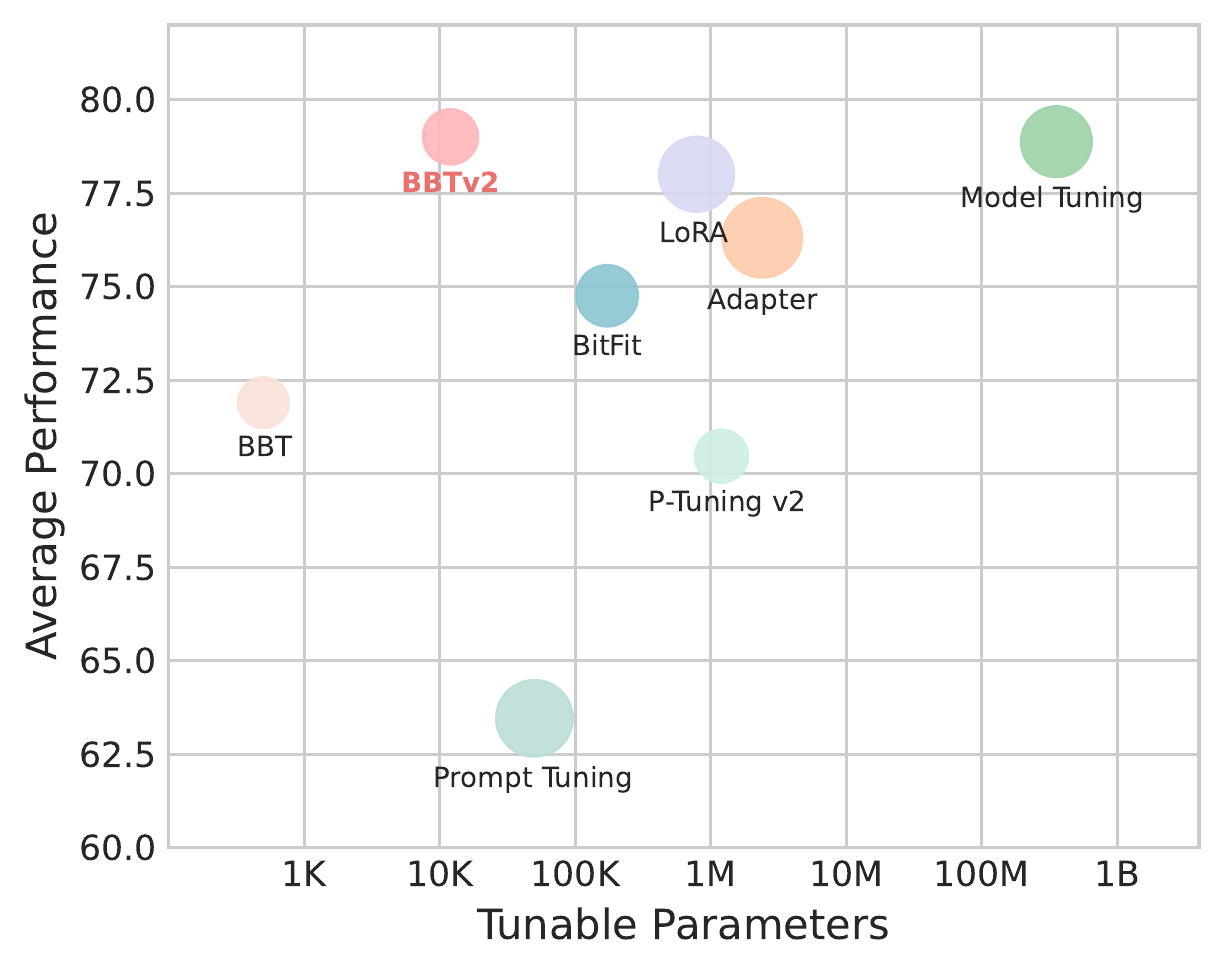}
    \caption{BBTv2 achieves comparable results to gradient-based methods on average performance over 7 language understanding tasks (§\ref{sec:data}) with much fewer tunable parameters. Size of the circle is proportional to the standard deviation of the performance. All the methods are evaluated on RoBERTa\textsubscript{LARGE}.}
    \label{fig:perf_overview}
\end{figure}

To make LLMs benefit a wider range of audiences, a common practice is to release LLMs as a service and allow users to access the powerful LLMs through their inference APIs~\cite{Brown2020GPT3}. In such a scenario, called Languaged-Model-as-a-Service (LMaaS)~\cite{Sun2022BBT}, users cannot access or tune model parameters but can only tune their prompts to accomplish language tasks of interest. \citet{Brown2020GPT3} propose to recast downstream tasks as a language modeling task and perform different tasks by conditioning on task-specific text prompts. Further, they demonstrate that LLMs exhibit an emergent ability of in-context learning, i.e., LLMs can learn to perform tasks with a few demonstrations provided in the input context without updating parameters. Nevertheless, its performance has been shown to highly depend on the choice of the prompt or the demonstrations~\cite{Zhao2021Calibrate,Liu2022What} and still lags far behind full model tuning. 

Recently, \citet{Sun2022BBT} proposed Black-Box Tuning (BBT), which optimizes continuous prompt by only accessing model inference APIs. In contrast to gradient-based tuning that requires expensive back-propagation\footnote{The computation and storage cost of back-propagation is proportional to the forward compute. More widely used variants of gradient descent, such as Adam~\cite{kingma2015adam}, even require higher compute resources.}, BBT only requires model forward computation, which can be highly optimized by acceleration frameworks such as ONNX Runetime and TensorRT. In addition, the optimization cost of BBT is decoupled from the scale of the model. Instead, larger models can be more favorable to BBT due to the lower intrinsic dimensionality~\cite{Aghajanyan2021Intrinsic}. Despite its efficiency superiority and comparable performance to gradient descent, our pilot experiments (§\ref{sec:pilot}) show that BBT lacks versatility across tasks and PTMs. 




In this paper, we present BBTv2, an improved version of BBT, to address these issues. Instead of injecting continuous prompt tokens merely in the input layer, BBTv2 prepends prompts to hidden states at every layer of the PTM (termed as \textit{deep prompts}), incorporating an order of magnitude more parameters to handle more difficult tasks. However, the increased number of parameters also poses a challenge for high-dimensional derivative-free optimization (DFO, \citet{Shahriari2016BO,Qian2021DFO}). Fortunately, we show that the forward computation of modern PTMs can be decomposed into an additive form w.r.t. hidden states of each layer thanks to the residual connections~\cite{He2016ResNet}. Hence, the optimization of the deep prompts can be decomposed into multiple low-dimensional sub-problems, each corresponding to the optimization of prompt at one layer. Based on this insight, we propose a divide-and-conquer algorithm to alternately optimize prompt at each layer.
For the optimization at each layer, we maintain a random linear transformation that projects the prompt parameters into a low-dimensional subspace and perform DFO in the generated subspace. To generalize BBTv2 to a variety of PTMs, we generate the random projections using normal distributions with PTM-related standard deviations.

Experimental results show that BBTv2 significantly improves BBT on average performance across 7 language understanding tasks. 
As shown in Figure~\ref{fig:perf_overview}, BBTv2 achieves comparable performance to full model tuning and state-of-the-art PET methods including Adapter~\cite{Houlsby2019Adapter}, BitFit~\cite{Zaken22BitFit}, LoRA~\cite{Hu2021LoRA}, and P-Tuning v2~\cite{Liu2021PTuningv2} while with much fewer tunable parameters. Code is publicly available at \url{https://github.com/txsun1997/Black-Box-Tuning}.

\section{Black-Box Tuning}
\label{sec:bbt}
Black-Box Tuning (BBT)~\cite{Sun2022BBT} is a derivative-free framework to drive PTMs for few-shot learning. In particular, for a batch of training data $(X,Y)$, we first convert the texts $X$ with some pre-defined templates (e.g., "\textit{It was} \texttt{[MASK]}") into $\Tilde{X}$, and the labels $Y$ with a pre-defined map into label words $\Tilde{Y}$ (e.g., "\textit{great}" and "\textit{terrible}"). By this, we can formulate various downstream tasks into a general-purpose (masked) language modeling task and utilize the pre-trained (masked) language modeling head to solve them. Assume the PTM inference API $f$ takes a continuous prompt $\mathbf{p}$ and a batch of converted texts $\Tilde{X}$ as input, and outputs the logits of the tokens of interest (e.g., the \texttt{[MASK]} token). BBT seeks to find the optimal prompt $\mathbf{p}^\star = \arg\min_{\mathbf{p}\in\mathcal{P}}\mathcal{L}(f(\mathbf{p},\Tilde{X}), \Tilde{Y})$, where $\mathcal{P}$ is the prompt space and $\mathcal{L}$ is some loss function such as cross entropy. The closed form and the gradients of $f$ are not accessible to BBT.

The prompt $\mathbf{p}\in\mathbb{R}^D$ usually has tens of thousands of dimensions, making it infeasible to be optimized with derivative-free optimization (DFO) algorithms. Hence, BBT adopts a random projection $\mathbf{A}\in\mathbb{R}^{D\times d}$ to generate a low-dimensional subspace $\mathcal{Z}\in\mathbb{R}^d$ and performs optimization in the generated subspace, i.e.,
\begin{align}
    \mathbf{z}^\star = \mathop{\arg\min}_{\mathbf{z}\in \mathcal{Z}}\mathcal{L}(f(\mathbf{Az}+\mathbf{p}_0,\Tilde{X}), \Tilde{Y}),\label{eq:bbt}
\end{align}
where $\mathbf{p}_0$ is the initial prompt embedding. If not using pre-trained prompt embedding, $\mathbf{p}_0$ is the word embeddings randomly drawn from the vocabulary. 

BBT adopts the Covariance Matrix Adaptation Evolution Strategy (CMA-ES)~\cite{Hansen2001CMA,Hansen2003Reducing} to optimize Eq.(\ref{eq:bbt}) and obtain the desired prompt $\mathbf{p}^\star=\mathbf{A}\mathbf{z}^\star$. The random projection $\mathbf{A}$ is frozen during optimization.

\section{Pilot Experiments}
\label{sec:pilot}

\subsection{Limitations of BBT Across Tasks}
\begin{figure}[t]
    \centering
    \includegraphics[width=\linewidth]{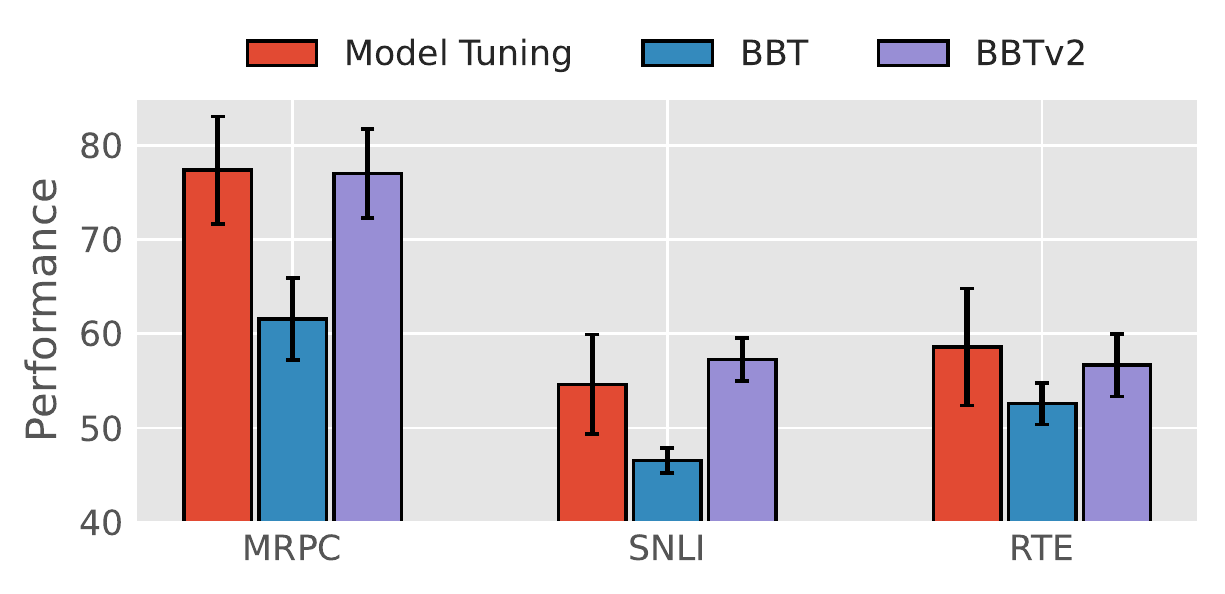}
    \caption{Performance on three entailment tasks. We report F1 score for MRPC and accuracy for SNLI and RTE. Without pre-trained prompt embedding, BBTv2 can still match the performance of full model tuning on entailment tasks under the 16-shot setting.}
    \label{fig:nli_perf}
\end{figure}

\paragraph{Unsatisfactory Performance on Entailment Tasks.}
As demonstrated by \citet{Sun2022BBT}, BBT can outperform model tuning on entailment tasks when using pre-trained prompt embedding for initialization. 
However, pre-trained prompt embedding is not always available for many languages and PTMs. 
Without pre-trained prompt embedding, BBT still lags far behind model tuning on entailment tasks. 
In other words, \textit{BBT does not completely get rid of the dependence on gradients to exceed model tuning}. 
In contrast, as depicted in Figure~\ref{fig:nli_perf}, BBTv2 can match the performance of model tuning on three entailment tasks, namely MRPC~\cite{Dolan2005MRPC}, SNLI~\cite{Bowman2015SNLI}, and RTE~\cite{Wang2019GLUE} without using pre-trained prompt embedding.

\begin{figure}[t]
    \centering
    \includegraphics[width=.9\linewidth]{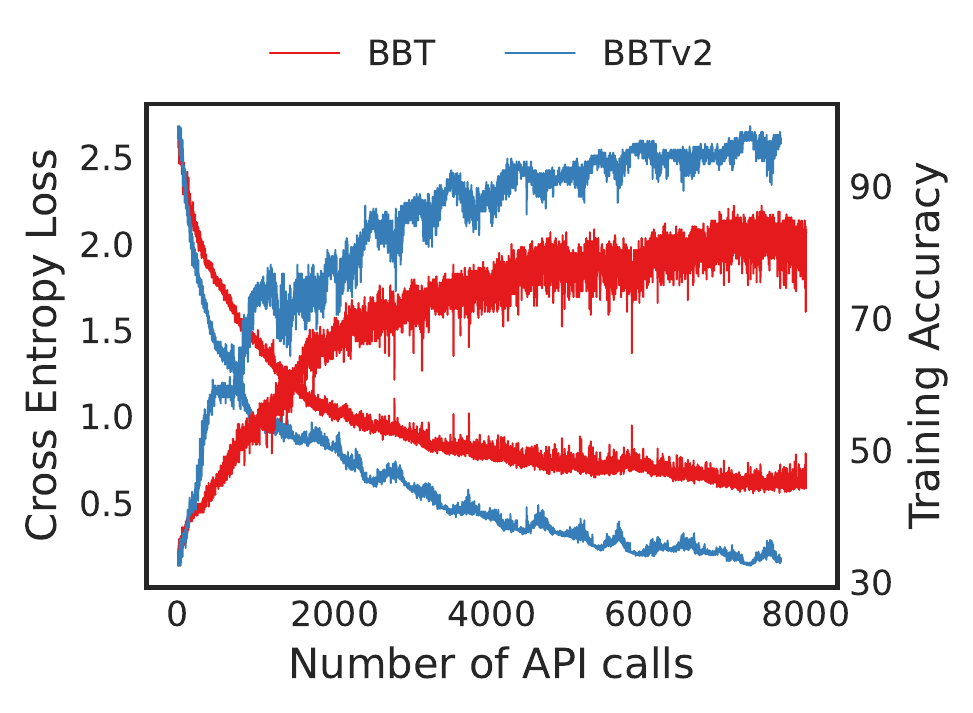}
    \caption{Comparison of the convergence rates of BBT and BBTv2 on DBPedia (14 classes).}
    \label{fig:dbpedia_convergence}
\end{figure}

\paragraph{Slow Convergence on Many-Label Classification Tasks.} 
BBT suffers from slow convergence rate when the number of labels becomes large.
As reported by \citet{Sun2022BBT}, BBT cannot converge within the budget of 8,000 API calls, which is sufficient for common tasks to converge, on DBPedia~\cite{Zhang2015Char}, a topic classification task with 14 labels.
Figure~\ref{fig:dbpedia_convergence} shows the cross entropy loss and the training accuracy during optimization. 
Compared with BBT, the proposed BBTv2 significantly accelerates convergence on DBPedia.

\subsection{Limitations of BBT Across PTMs}
\label{sec:limit_across_models}
\begin{figure}[t]
    \centering
    \begin{subfigure}{\linewidth}
    \centering
    \includegraphics[width=\linewidth]{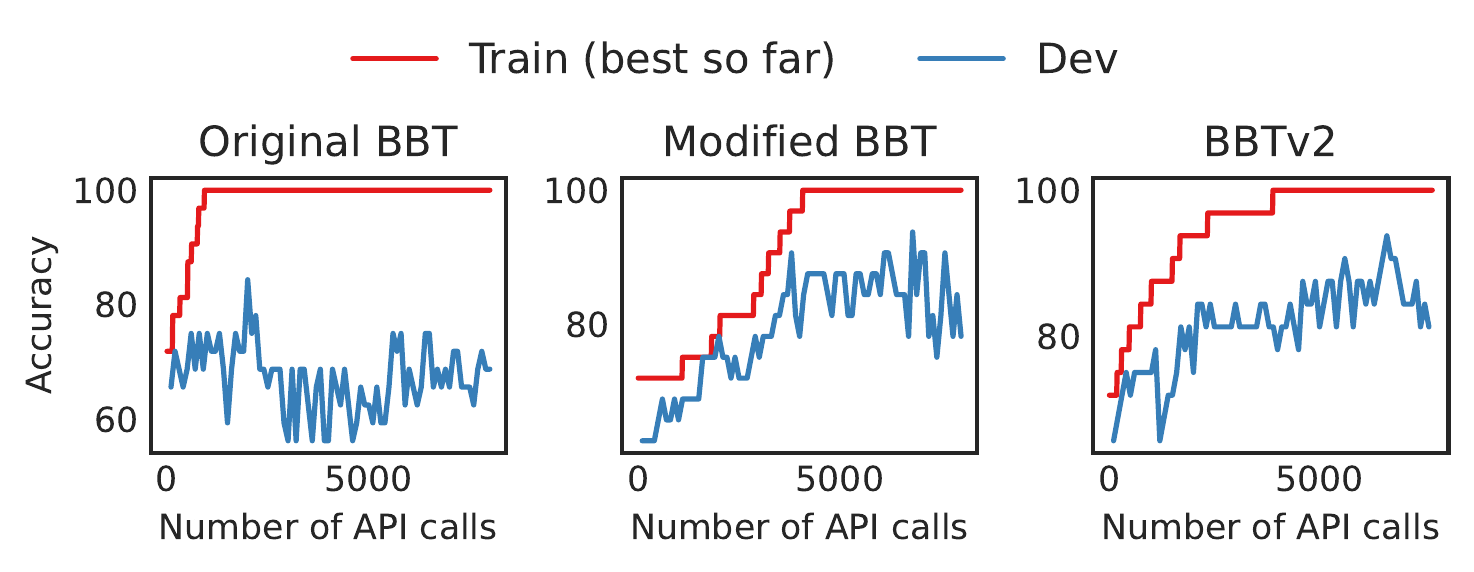}
    \caption{BERT\textsubscript{LARGE}}
    \end{subfigure}
    \begin{subfigure}{\linewidth}
    \centering
    \includegraphics[width=\linewidth]{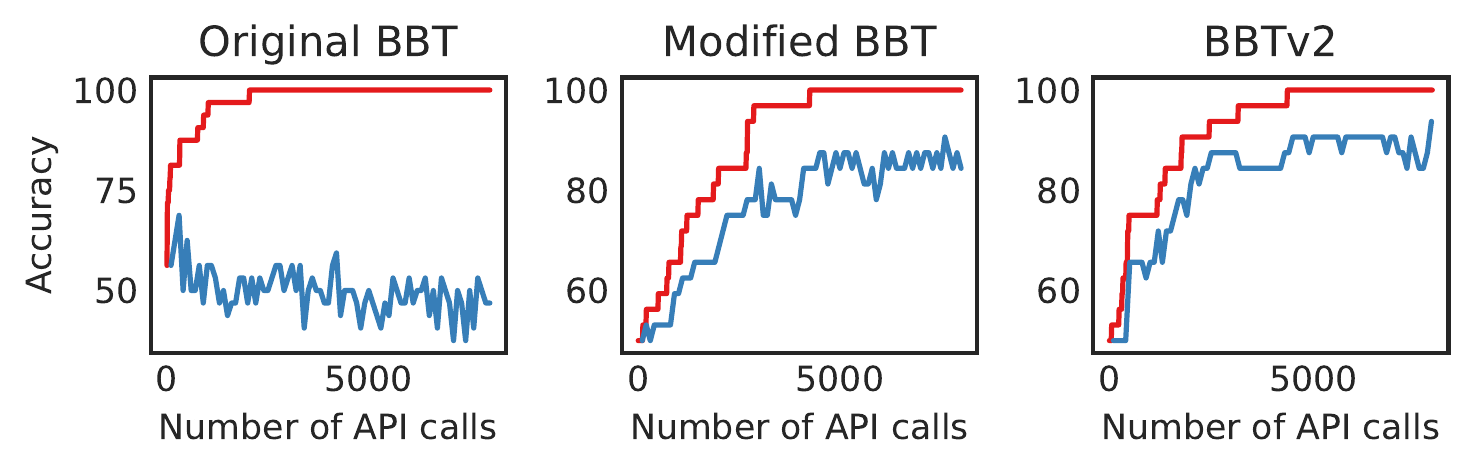}
    \caption{BART\textsubscript{LARGE}}
    \end{subfigure}
    \caption{Accuracy on the 16-shot training and development set of SST-2 with BERT\textsubscript{LARGE} and BART\textsubscript{LARGE}. The original BBT tends to overfit training data. By using normal distributions with the standard deviations calculated by Eq.(\ref{eq:sigma}) to generate random projections, the modified BBT and BBTv2 can generalize well to development sets.}
    \label{fig:overfit}
\end{figure}

\paragraph{Overfitting on Training Data.}
When switching the backbone model from RoBERTa~\cite{Liu2019roberta} to other PTMs, we find that BBT tends to overfit training data. As shown in Figure~\ref{fig:overfit}, the original BBT with BERT\textsubscript{LARGE}~\cite{Devlin2019BERT} and BART\textsubscript{LARGE}~\cite{Lewis2020BART} can achieve 100\% accuracy on the SST-2 training set, but achieves little improvement on the development set. We conjecture that the random projection adopted by the original BBT hinders its generalization. By generating random projections using normal distributions with model-related standard deviations (§\ref{sec:normal_distribution}), our modified BBT and BBTv2 exhibit stronger generalization ability.

\section{BBTv2}

\begin{figure*}
    \centering
    \includegraphics[width=\linewidth]{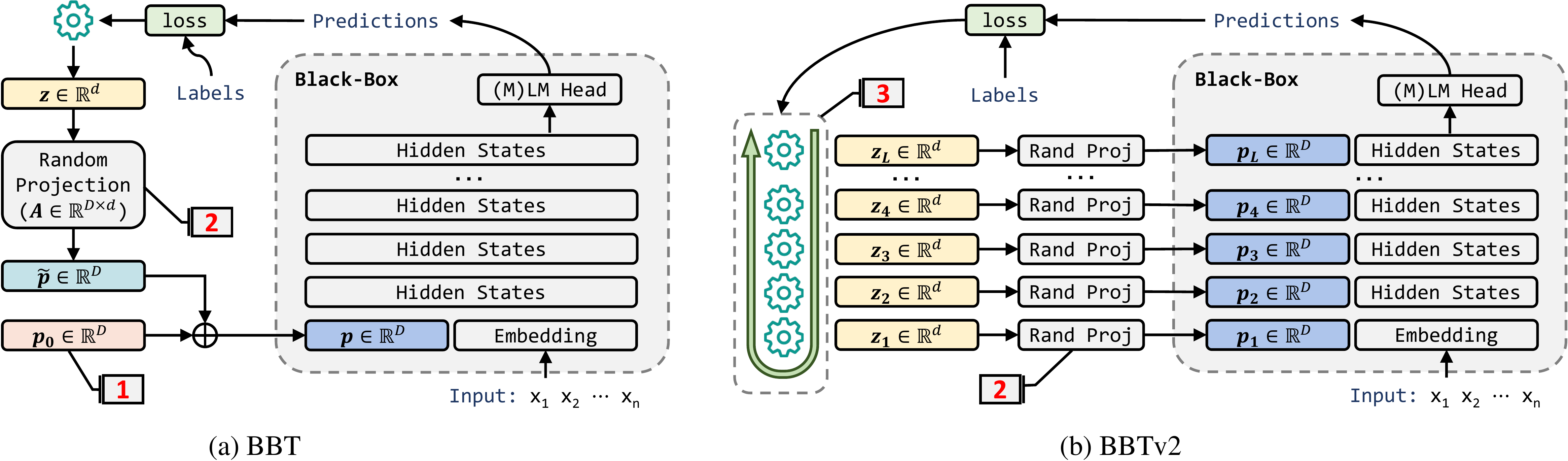}
    \caption{An illustration of (a) BBT~\cite{Sun2022BBT} and (b) BBTv2. \includegraphics[width=.28cm]{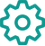}\ \ is some derivative-free optimizer such as CMA-ES. Compared with BBT, BBTv2 has 3 differences: (1) BBT requires pre-trained prompt embedding $\mathbf{p}_0$ to match the performance of model tuning on entailment tasks, and therefore does not completely get rid of gradients. In contrast, BBTv2 requires no prompt pre-training. (2) BBT generates the random projection using a uniform distribution while BBTv2 adopts model-specific normal distributions. (3) Instead of optimizing the prompt merely in the input layer, BBTv2 uses a divide-and-conquer algorithm to alternately optimize prompt at each layer.}
    \label{fig:bbtv2}
\end{figure*}

\begin{algorithm}[t]
	\caption{DC Algorithm for BBTv2}
	\label{alg:deepbbt}
	\begin{algorithmic}[1]
		\REQUIRE $L$-layer PTM Inference API $f$,\\
		\ \ \ \ \ \ \ \ \ \ Loss function $\mathcal{L}$,\\
		\ \ \ \ \ \ \ \ \ \ Budget of API calls $\mathcal{B}$,\\
		\ \ \ \ \ \ \ \ \ \ Derivative-free optimizers $\{\mathcal{M}_j\}_{j=1}^L$
		\STATE Initialize random projections $\mathbf{A}_1, \dots, \mathbf{A}_L$ 
		\STATE Initialize parameters $\mathbf{z}_1^{(0)}, \dots, \mathbf{z}_L^{(0)}$
		\STATE Deep prompts $\mathbf{p} = \langle \mathbf{A}_1\mathbf{z}_1^{(0)}, \dots, \mathbf{A}_L\mathbf{z}_L^{(0)}\rangle$
		\FOR {$i=1$ to $\mathcal{B}/L$}
		    \FOR {$j=1$ to $L$}
		        \STATE Evaluate: $loss = \mathcal{L}(f(\mathbf{p}))$
		        \STATE Update: $\mathbf{z}_j^{(i)} \gets \mathcal{M}_j(\mathbf{z}_j^{(i-1)}, loss)$
		        \STATE Replace: $\mathbf{p}_j \gets \mathbf{A}_j\mathbf{z}_j^{(i)}$
		    \ENDFOR
		\ENDFOR
		\STATE \textbf{return} Optimized deep prompts $\mathbf{p}$
	\end{algorithmic}
\end{algorithm}

\subsection{Deep Black-Box Tuning}
\label{sec:deepbbt}
Though BBT has achieved comparable performance to model tuning on simple classification tasks (e.g., SST-2), our pilot experiments (§\ref{sec:pilot}) show that it lacks versatility across tasks and PTMs. As an improved variant of BBT, BBTv2 seeks to generalize BBT across tasks and PTMs.

Inspired by the recent success of \textit{deep prompt tuning}~\citep{Li2021Prefix,Qin21Learning,Liu2021PTuningv2}, we manage to inject continuous prompt tokens to every layer of the PTM and optimize them with derivative-free methods. Compared to BBT that optimizes the prompt merely in the input layer, BBTv2 has an order of magnitude more parameters. For a PTM with $L$ layers, BBTv2 seeks to optimize $\mathbf{p}=\langle\mathbf{p}_1,\dots,\mathbf{p}_L\rangle$, where $\mathbf{p}_i\in\mathbb{R}^D$. Hence, the number of parameters to be optimized becomes $LD$. Say we are using RoBERTa\textsubscript{LARGE} with 24 layers and insert 50 prompt tokens at each layer, the total number of parameters to be optimized is 1.2M, posing a challenge of high-dimensional DFO. Instead of simply extending the dimensionality of the random projection matrix $\mathbf{A}$ to $\mathbb{R}^{LD\times d}$~\footnote{In our preliminary experiments, we implement this version but did not obtain positive results.}, we propose a divide-and-conquer (DC) algorithm to handle the increased parameters.

In fact, DC has been well explored in prior work~\cite{Kandasamy2015HighDim,Mei2016DC} to cope with high-dimensional DFO problems by decomposing the original high-dimensional problem into multiple low-dimensional sub-problems, and solving them separately. 
The key assumption of applying DC is that, the objective $f$ can be decomposed into some additive form. 
Fortunately, the forward computation of modern PTMs can be expanded into an additive form due to the residual connections~\cite{He2016ResNet}. For instance, the forward computation of a three-layered PTM can be decomposed as
\begin{align}
    f(\mathbf{x}_1) &= f_3(\mathbf{x}_3) + \mathbf{x}_3\\
    &= f_3(\mathbf{x}_3) + f_2(\mathbf{x}_2) + \mathbf{x}_2\\
    &= f_3(\mathbf{x}_3) + f_2(\mathbf{x}_2) + f_1(\mathbf{x}_1) + \mathbf{x}_1,\label{eq:decompose}
\end{align}
where $f_i$ is the transformation function of the $i$-th layer, $\mathbf{x}_i$ is the input of the $i$-th layer, and $\mathbf{x}_1$ is the input embedding. Thus, optimizing the continuous prompts $\{\mathbf{p}_i\}_{i=1}^L$ attached to the hidden states at every layer $\{\mathbf{x}_i\}_{i=1}^L$ can be regarded as independent sub-problems.\footnote{We omit the classification head on the top of the PTM since it is usually a linear transformation and would not affect the additive decomposition.} 
Since the assumption is satisfied, we propose a DC-based algorithm, which is described in Algorithm~\ref{alg:deepbbt}, to implement BBTv2.

The prompts at different layers are optimized alternately from bottom to up. For the optimization at each layer, we maintain a specific random projection $\mathbf{A}_j$ and a CMA-ES optimizer $\mathcal{M}_j$. When alternating to layer $j$ (Line 6-8 in Algorithm~\ref{alg:deepbbt}), a single CMA-ES iteration is performed in the same fashion as BBT, i.e., a new $\mathbf{z}_j$ is generated by $\mathcal{M}_j$ and is then projected to $\mathbf{p}_j$ using $\mathbf{A}_j$. A graphical illustration is shown in Figure~\ref{fig:bbtv2}.

During PTM inference, $\mathbf{p}_j=\mathbf{A}_j\mathbf{z}_j$ is first added with an initial prompt embedding $\mathbf{p}_{j}^0$ and then concatenated with the hidden states $\mathbf{x}_j$. Thus, according to Eq.(\ref{eq:decompose}), the forward computation of a $L$-layered PTM can be viewed as
\begin{align}
    f(\mathbf{x}_1,\mathbf{p}_{1:L}) =& [\mathbf{A}_1{\color{RedOrange}{\mathbf{z}_1}}+\mathbf{p}_1^0;\mathbf{x}_1]\nonumber\\ &+\sum_{j=1}^{L}f_j([\mathbf{A}_j{\color{RedOrange}{\mathbf{z}_j}}+\mathbf{p}_j^0;\mathbf{x}_j]),
\end{align}
where $[\cdot ;\cdot]$ means concatenation. Tunable parameters are highlighted in {\color{RedOrange}{color}}. We set $\mathbf{p}_1^0$ as the word embeddings randomly drawn from the PTM vocabulary for all tasks. $\mathbf{p}_j^0$ ($1<j\leq L$) is the hidden states of the prompt tokens at the $j$-th layer. 

\subsection{Revisiting Random Projection}
\label{sec:normal_distribution}
In the original BBT, each entry in the random projection $\mathbf{A}$ is sampled from a uniform distribution~\cite{He2015Delving}. In their experiments, using normal distribution $\mathcal{N}(0, 1/d)$ to generate the random projection results in slow convergence and inferior performance. However, we show in pilot experiments that the uniform distribution exhibits poor generalization on PTMs other than RoBERTa. In this section, we shed some light on the effect of the random projection, and propose to use normal distributions with model-related standard deviations to generate random projections. In fact, most prior works in high-dimensional DFO~\cite{Wang2016BORE,Qian2016DFOHigh} also adopt normal distributions to generate random projections. However, they usually simply use $\mathcal{N}(0,1)$ or $\mathcal{N}(0, 1/d)$, both of which underperform in our scenario.

\begin{figure}[t]
    \centering
    \includegraphics[width=\linewidth]{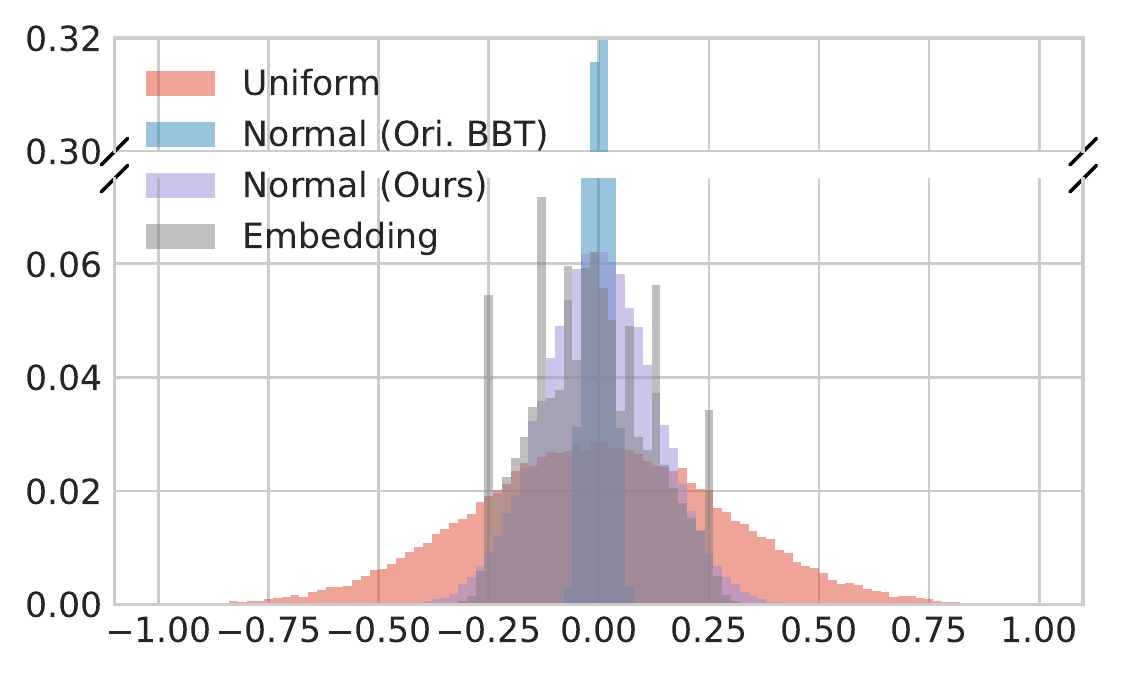}
    \caption{Distributions of RoBERTa word embeddings and generated prompt $\mathbf{p}=\mathbf{Az}$ where $\mathbf{A}$ is sampled from different distributions. When using our designed normal distribution to generate the random projection $\mathbf{A}$, the distribution of the projected prompt well matches the shape of the word embedding distribution, leading to faster convergence and stronger generalization.}
    \label{fig:distributions}
\end{figure}

To take a closer look into the effect of the random projection, we draw distribution of the initial prompt $\mathbf{p}$ that is projected from $\mathbf{z}$ by the projection matrix $\mathbf{A}$. Here, $\mathbf{z}$ is sampled from the normal distribution maintained by the CMA-ES, which is initially set to $\mathcal{N}(0,0.5)$ in BBT. By generating $\mathbf{A}$ from different distributions, we simulate the distribution of the projected prompt $\mathbf{p}$ and compare with the distribution of RoBERTa\textsubscript{LARGE} word embeddings.\footnote{We hypothesis that a high-quality prompt $\mathbf{p}$ should lie within the distribution of word embeddings (hidden states).} As revealed by Figure~\ref{fig:distributions}, when $\mathbf{A}$ is sampled from the normal distribution used in the original BBT, i.e., $\mathcal{N}(0, 1/d)$, the projected prompt $\mathbf{p}$ cannot cover the range of word embeddings, and therefore suffers from slow convergence. In contrast, using uniform distribution can cover the range of word embeddings, which explains why it performs well on RoBERTa\textsubscript{LARGE}.

Thus, to generalize BBT (and BBTv2) across different PTMs, we have to take into account the distribution of word embeddings (and hidden states for BBTv2) of the PTM for generating random projections. In particular, we use the normal distribution with standard deviation as follows,
\begin{align}
    \sigma_A = \frac{\alpha\hat{\sigma}}{\sqrt{d}\sigma_z},\label{eq:sigma}
\end{align}
where $\hat{\sigma}$ is observed standard deviation of word embeddings (or hidden states for BBTv2), $\sigma_z$ is the standard deviation of the normal distribution maintained by CMA-ES, $\alpha$ is a constant scalar to stretch the distribution. Initially, we set $\mu_z=\mu_A=0$ so that no prior knowledge about the optimization direction is incorporated. The main idea behind the above calculation is to match the distribution (more specifically the variance) between the projected prompt and word embeddings (or hidden states). When $\alpha=1$, as we can see in Figure~~\ref{fig:distributions}, the distribution of the projected prompt can perfectly match the distribution of the word embeddings. Detailed derivation of Eq.(\ref{eq:sigma}) is provided in Appendix~\ref{sec:append_A}.

\section{Experiments}
\begin{table*}[t!]
\centering
\resizebox{\linewidth}{!}{
\begin{tabular}{lccccccccr}
\toprule
\multirow{2}{*}{\textbf{Method}} & \textbf{Tunable} & \textbf{SST-2} & \textbf{Yelp P.} & \textbf{AG's News} & \textbf{DBPedia} & \textbf{MRPC} & \textbf{SNLI} & \textbf{RTE} & \multirow{2}{*}{\textbf{Avg.}} \\
& \textbf{Params} & acc & acc & acc & acc & F1 & acc & acc & \\ \midrule
\multicolumn{9}{c}{\textit{Gradient-Based Methods}}                                                                                                                                                                      \\ \midrule
Model Tuning & \cellcolor[gray]{.5}355M & \underline{85.39} \small{$\pm$2.84} & \underline{91.82} \small{$\pm$0.79} & 86.36 \small{$\pm$1.85} & \underline{97.98} \small{$\pm$0.14} & \textbf{77.35} \small{$\pm$5.70} & 54.64 \small{$\pm$5.29} & \textbf{58.60} \small{$\pm$6.21} & \textbf{78.88}\\
Adapter    & \cellcolor[gray]{.63}2.4M & 83.91 \small{$\pm$2.90} & 90.99 \small{$\pm$2.86}  & 86.01 \small{$\pm$2.18} & \textbf{97.99} \small{$\pm$0.07} & 69.20 \small{$\pm$3.58} & \underline{57.46} \small{$\pm$6.63} & 48.62 \small{$\pm$4.74} & 76.31 \\ 
BitFit    & \cellcolor[gray]{.7}172K & 81.19 \small{$\pm$6.08} & 88.63 \small{$\pm$6.69} & \underline{86.83} \small{$\pm$0.62} & 94.42 \small{$\pm$0.94} & 66.26 \small{$\pm$6.81} & 53.42 \small{$\pm$10.63} & 52.59 \small{$\pm$5.31} & 74.76 \\ 
LoRA    & \cellcolor[gray]{.66}786K & \textbf{88.49} \small{$\pm$2.90} & 90.21 \small{$\pm$4.00} & \textbf{87.09} \small{$\pm$0.85} & 97.86 \small{$\pm$0.17} & \underline{72.14} \small{$\pm$2.23} & \textbf{61.03} \small{$\pm$8.55} & 49.22 \small{$\pm$5.12} & \underline{78.01} \\ 
Prompt Tuning & \cellcolor[gray]{.73}50K & 68.23 \small{$\pm$3.78} & 61.02 \small{$\pm$6.65} & 84.81 \small{$\pm$0.66} & 87.75 \small{$\pm$1.48} & 51.61 \small{$\pm$8.67} & 36.13 \small{$\pm$1.51} & \underline{54.69} \small{$\pm$3.79} & 63.46 \\
P-Tuning v2    & \cellcolor[gray]{.65}1.2M &64.33 \small{$\pm$3.05} &\textbf{92.63} \small{$\pm$1.39} &83.46 \small{$\pm$1.01} &97.05 \small{$\pm$0.41} &68.14 \small{$\pm$3.89} &36.89 \small{$\pm$0.79} &50.78 \small{$\pm$2.28} & 70.47 \\ 
\midrule
\multicolumn{9}{c}{\textit{Gradient-Free Methods}}                                                                                                                                                                      \\ \midrule
Manual Prompt            & 0 &  79.82 & 89.65 & 76.96 & 41.33 & 67.40 & 31.11 & 51.62 & 62.56 \\
In-Context Learning      & 0 & 79.79 \small{$\pm$3.06} & 85.38 \small{$\pm$3.92} & 62.21 \small{$\pm$13.46} & 34.83 \small{$\pm$7.59} & 45.81 \small{$\pm$6.67} & \underline{47.11} \small{$\pm$0.63} & \textbf{60.36} \small{$\pm$1.56} & 59.36\\
Feature-MLP           & \cellcolor[gray]{.65}1M & 64.80 \small{$\pm$1.78} &79.20 \small{$\pm$2.26} &70.77 \small{$\pm$0.67} & 87.78 \small{$\pm$0.61}&68.40 \small{$\pm$0.86} & 42.01 \small{$\pm$0.33} &53.43 \small{$\pm$1.57} & 66.63\\
Feature-BiLSTM           & \cellcolor[gray]{.58}17M & 65.95 \small{$\pm$0.99} &74.68 \small{$\pm$0.10} &77.28 \small{$\pm$2.83} & \underline{90.37} \small{$\pm$3.10} & \underline{71.55} \small{$\pm$7.10} & 46.02 \small{$\pm$0.38} &52.17 \small{$\pm$0.25} & 68.29\\
BBT & \cellcolor[gray]{.85}500 & \underline{89.56} \small{$\pm$0.25} & \underline{91.50} \small{$\pm$0.16} & \underline{81.51} \small{$\pm$0.79} & 79.99$^\star$\small{$\pm$2.95} & 61.56 \small{$\pm$4.34} &  46.58 \small{$\pm$1.33}  & 52.59 \small{$\pm$2.21} & \underline{71.90} \\ 
\textbf{BBTv2}    & \cellcolor[gray]{.77}12K & \textbf{90.33} \small{$\pm$1.73} & \textbf{92.86} \small{$\pm$0.62} & \textbf{85.28} \small{$\pm$0.49} & \textbf{93.64} \small{$\pm$0.68} & \textbf{77.01} \small{$\pm$4.73} & \textbf{57.27} \small{$\pm$2.27} & \underline{56.68} \small{$\pm$3.32} & \textbf{79.01} \\ 
\bottomrule
\end{tabular}
}
\caption{Overall comparison on various language understanding tasks. We report mean and standard deviation of performance over 3 different splits (§\ref{sec:data}). All of the results are obtained with pre-trained RoBERTa\textsubscript{LARGE} in the 16-shot (per class) setting. In each track, the best results are highlighted in \textbf{bold} and the second best results are marked with \underline{underline}. $^\star$ We reimplement BBT on DBPedia given a budget of 8,000 API calls for fair comparison.}
\label{tab:main_results}
\end{table*}

\subsection{Datasets and Tasks}
\label{sec:data}
For comparison, we evaluate BBTv2 on the same datasets as BBT, i.e., SST-2~\cite{Socher2013SST}, Yelp~\cite{Zhang2015Char}, AG's News~\cite{Zhang2015Char}, DBPedia~\cite{Zhang2015Char}, SNLI~\cite{Bowman2015SNLI}, RTE~\cite{Wang2019GLUE}, and MRPC~\cite{Dolan2005MRPC}. SST-2 and Yelp are sentiment analysis tasks, AG's News and DBPedia are topic classification tasks, SNLI and RTE are natural language inference (NLI) tasks, and MRPC is a paraphrase task. In addition, we include two Chinese tasks, ChnSent\footnote{\url{https://github.com/SophonPlus/ChineseNlpCorpus}} and LCQMC~\cite{Liu2018LCQMC}, for evaluation on CPM-2~\cite{Zhang2021CPM2}, a Chinese PTM with $\sim$11B parameters. ChnSent is a sentiment analysis task while LCQMC is a question matching task.

We follow the same procedure as \citet{Zhang2021Revisiting,Gu2021PPT,Sun2022BBT} to construct the true few-shot learning settings~\cite{Perez2021TrueFewShot}. 
In particular, we
randomly draw $k$ samples for each class to construct a $k$-shot training set $\mathcal{D}_{\text{train}}$, and construct a development set $\mathcal{D}_{\text{dev}}$ by randomly selecting another $k$ samples from the original training set such that $|\mathcal{D}_{\text{train}}|=|\mathcal{D}_{\text{dev}}|$. We use the original development sets as the test sets. For datasets without development sets, we use the original test sets. Therefore, in our experiments we have $|\mathcal{D}_{\text{test}}| \gg |\mathcal{D}_{\text{train}}| = |\mathcal{D}_{\text{dev}}|$.

\subsection{Baselines}
We consider two types of methods as our baselines: \textit{gradient-based methods} and \textit{gradient-free methods}.

For gradient-based methods, we compare with \textbf{(1) Model Tuning} and state-of-the-art parameter-efficient methods including \textbf{(2) Adapter}~\cite{Houlsby2019Adapter}, \textbf{(3) BitFit}~\cite{Zaken22BitFit}, \textbf{(4) LoRA}~\cite{Hu2021LoRA}, \textbf{(5) Prompt Tuning}~\cite{Lester2021Prompt}, and \textbf{(6) P-Tuning v2}~\cite{Liu2021PTuningv2}. We implement Adapter, BitFit, and LoRA using OpenDelta\footnote{\url{https://github.com/thunlp/OpenDelta}}, and evaluate P-Tuning v2 in our experimental settings based on the official implementation\footnote{\url{https://github.com/THUDM/P-tuning-v2}}. The results of Model Tuning and Prompt Tuning are taken from \citet{Sun2022BBT}.

For gradient-free methods, we compare with two non-learning prompt-based methods: \textbf{(1) Manual Prompt} and \textbf{(2) In-Context Learning}~\cite{Brown2020GPT3}; two feature-based methods: \textbf{(3) Feature-MLP} and \textbf{(4) Feature-BiLSTM}, which is to train a MLP/BiLSTM classifier on the features extracted by the PTM; and \textbf{(5) BBT}~\cite{Sun2022BBT}. The results of these gradient-free baselines are taken from \citet{Sun2022BBT}. One exception is the performance of BBT on DBPedia. In the original paper, BBT is performed given a larger budget (20,000 API calls) on DBPedia for convergence. In this work, we reimplement BBT on DBPedia with the same budget (8,000 API calls) as other tasks for fair comparison.

\subsection{Implementation Details}
\paragraph{Backbones.}
To compare with BBT, we mainly use RoBERTa\textsubscript{LARGE}~\cite{Liu2019roberta} as our backbone model. To verify the versatility, we also evaluate on other PTMs including BERT\textsubscript{LARGE}~\cite{Devlin2019BERT}, GPT-2\textsubscript{LARGE}, BART\textsubscript{LARGE}~\cite{Lewis2020BART}, and T5\textsubscript{LARGE}~\cite{Raffel2020T5}. In addition, we also evaluate BBTv2 on a supersized Chinese PTM, CPM-2~\cite{Zhang2021CPM2}, which has $\sim$11B parameters.

\paragraph{Hyperparameters.} Most of the hyperparameters remain the same as BBT. We insert 50 continuous prompt tokens at each layer. The subspace dimensionality is set to 500. The CMA-ES with population size of 20 and budget of 8,000 API calls is applied to all the tasks. We adopt cross entropy as the loss function. For generating random projections, we use normal distributions with standard deviations calculated by Eq.(\ref{eq:sigma}) instead of uniform distributions.

\begin{table}[t]
\centering
\resizebox{\linewidth}{!}{
\begin{tabular}{l|cc|cc}
\toprule
                           & \multicolumn{2}{c|}{\textbf{SST-2}} & \multicolumn{2}{c}{\textbf{AG's News}} \\
                           & \multicolumn{2}{c|}{(max seq len: 47)}      & \multicolumn{2}{c}{(max seq len: 107)}         \\ \midrule
                           & \textbf{BBT}   & \textbf{BBTv2}  & \textbf{BBT}    & \textbf{BBTv2}    \\ \midrule
\textbf{Accuracy}          & 89.4           & 91.4               & 82.6            & 85.5                 \\ \midrule
\textbf{Training Time} &                &                    &                 &                      \\
\ PyTorch (mins)                    & 14.8           & 11.0               & 28.3            & 25.0                 \\
\ ONNX (mins)                      & 6.1            & 4.6                & 17.7            & 10.4                 \\ \midrule
\textbf{Memory}            &                &                    &                 &                      \\
\ User (MB)                  & 30             & 143                & 30              & 143                  \\
\ Server (GB)                & 3.0            & 3.0                & 4.6             & 4.6                  \\ \midrule
\textbf{Network}      &                &                    &                 &                      \\
\ Upload (KB)                     & 6              & 52                & 22              & 68                  \\
\ Download (KB)                   & 0.25           & 0.25               & 1               & 1                    \\ \bottomrule
\end{tabular}
}
\caption{Comparison of BBT and BBTv2 on accuracy, training time, memory use, and network load.}
\label{fig:detailed_comp}
\end{table}

\subsection{Results}
\paragraph{Overall Comparison.}
As shown in Table~\ref{tab:main_results}, BBTv2 outperforms BBT and other gradient-free methods on 6/7 tasks. In contrast to BBT, the improvement of BBTv2 mainly comes from DBPedia, which has 14 classes, and hard entailment tasks, namely MRPC and SNLI. On simple tasks such as SST-2 and Yelp, BBT can perform on par with BBTv2. When compared with gradient-based methods, BBTv2 achieves the best result in average across the 7 tasks while maintaining much fewer tunable parameters. It is worth noting that BBTv2, without any gradient-based components (e.g., the pre-trained prompt embedding used in BBT on entailment tasks~\cite{Sun2022BBT} or the white-box prompt optimization required by \citet{Diao2022Black}), is \textit{the first pure black-box method that matches the performance of full model tuning on various understanding tasks. }

\paragraph{Detailed Comparison.}
In Table~\ref{fig:detailed_comp}, we compare BBTv2 with BBT in other dimensions than accuracy. In addition to the improvement in accuracy, BBTv2 also confers faster convergence than BBT. For fair comparison of training time, we perform early stopping if the development accuracy does not increase after 1,000 steps. We report training times under two implementations, PyTorch~\cite{Paszke2019Pytorch} and ONNX Runtime\footnote{\url{https://onnxruntime.ai/}}, on a single NVIDIA GTX 3090 GPU. In terms of memory footprint and network load, BBTv2 slightly increases the memory use on the user side and the amount of data to be uploaded. 

\paragraph{BBTv2 Across PTMs.}
To verify the universality of BBTv2 across PTMs, we also evaluate on BERT, GPT-2, BART, T5, and CPM-2. As shown in Table~\ref{tab:other_ptm}, BBTv2 achieves superior performance over BBT on PTMs with varying architectures, i.e., encoder-only, decoder-only, and encoder-decoder PTMs.\footnote{We use the normal distribution (Eq.(\ref{eq:sigma})) for BBT for generalizing to PTMs other than RoBERTa.} In addition, we also verify the effectiveness of BBTv2 on a supersized Chinese PTM, CPM-2, which has $\sim$11B parameters. As shown in Table~\ref{tab:cpm}, when using CPM-2 as the backbone, BBTv2 outperforms full model tuning on two Chinese tasks. The results of Vanilla PT, Hybrid PT, and LM Adaption, which are three variants of prompt tuning, are taken from \citet{Gu2021PPT}. 

\begin{table}[t]
\centering
\resizebox{\linewidth}{!}{
\begin{tabular}{lcccc}
\toprule
\textbf{LM}              & \textbf{Method} & \textbf{SST-2} & \textbf{AG's News} & \textbf{DBPedia} \\ \midrule
\multicolumn{5}{c}{\textit{Encoder-only PTMs}}                                                           \\ \midrule
\multirow{2}{*}{BERT}    & BBT             & 76.26 \small{$\pm$2.64}   & 76.67 \small{$\pm$1.12}       & 89.58 \small{$\pm$0.51}     \\
                         & BBTv2           & 79.32 \small{$\pm$0.29}   & 79.58 \small{$\pm$1.15}       & 93.74 \small{$\pm$0.50}     \\
\multirow{2}{*}{RoBERTa} & BBT             & 89.56 \small{$\pm$0.25}   & 81.51 \small{$\pm$0.79}       & 79.99 \small{$\pm$2.95}     \\
                         & BBTv2           & 90.33 \small{$\pm$1.73}   & 85.28 \small{$\pm$0.49}       & 93.64 \small{$\pm$0.68}     \\ \midrule
\multicolumn{5}{c}{\textit{Decoder-only PTMs}}                                                           \\ \midrule
\multirow{2}{*}{GPT-2}   & BBT             & 75.53 \small{$\pm$1.98}   & 77.63 \small{$\pm$1.89} & 77.46 \small{$\pm$0.69} \\
                         & BBTv2           & 83.72 \small{$\pm$3.05}   & 79.96 \small{$\pm$0.75} & 91.36 \small{$\pm$0.73} \\ \midrule
\multicolumn{5}{c}{\textit{Encoder-Decoder PTMs}}                                                        \\ \midrule
\multirow{2}{*}{BART}    & BBT             & 77.87 \small{$\pm$2.57}   & 77.70 \small{$\pm$2.46}       & 79.64 \small{$\pm$1.55}     \\
                         & BBTv2           & 89.53 \small{$\pm$2.02}   & 81.30 \small{$\pm$2.58}       & 87.10 \small{$\pm$2.01}     \\
\multirow{2}{*}{T5}      & BBT             & 89.15 \small{$\pm$2.01}   & 83.98 \small{$\pm$1.87}       & 92.76 \small{$\pm$0.83}     \\
                         & BBTv2           & 91.40 \small{$\pm$1.17}   & 85.11 \small{$\pm$1.11}       & 93.36 \small{$\pm$0.80}     \\ \bottomrule
\end{tabular}
}
\caption{Comparison of BBT and BBTv2 on the large versions of BERT, RoBERTa, GPT-2, BART and T5.}
\label{tab:other_ptm}
\end{table}


\begin{table}[t]
\centering
\resizebox{\linewidth}{!}{
\begin{tabular}{lccc}
\toprule
\multirow{2}{*}{\textbf{Method}}  & \textbf{Tunable} & \textbf{ChnSent} & \textbf{LCQMC}      \\
& \textbf{Params} & acc & acc \\\midrule
Model Tuning& 11B & \underline{86.1} \small{$\pm$1.8}  & \underline{58.8} \small{$\pm$1.8} \\
Vanilla PT  & 410K & 62.1 \small{$\pm$3.1} & 51.5 \small{$\pm$3.4} \\
Hybrid PT   & 410K & 79.2 \small{$\pm$4.0} & 54.6 \small{$\pm$2.3} \\
LM Adaption & 410K & 74.3 \small{$\pm$5.2} & 51.4 \small{$\pm$2.9} \\\midrule
\textbf{BBTv2}    & 4.8K & \textbf{86.4} \small{$\pm$0.8} & \textbf{59.1} \small{$\pm$2.5}\\\bottomrule
\end{tabular}
}
\caption{Results on two Chinese tasks with CPM-2 as the backbone PTM.}
\label{tab:cpm}
\end{table}

\subsection{Ablations}
\label{sec:ablation}
\begin{figure}[t]
    \centering
    \includegraphics[width=.49\linewidth]{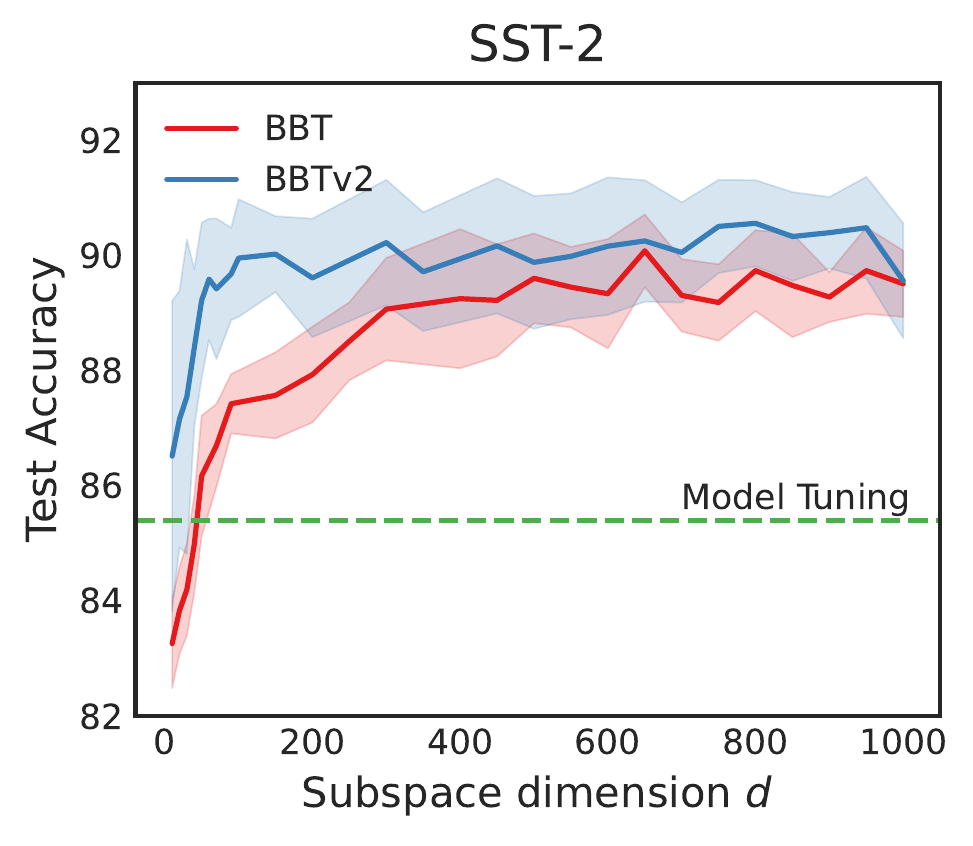}
    \includegraphics[width=.49\linewidth]{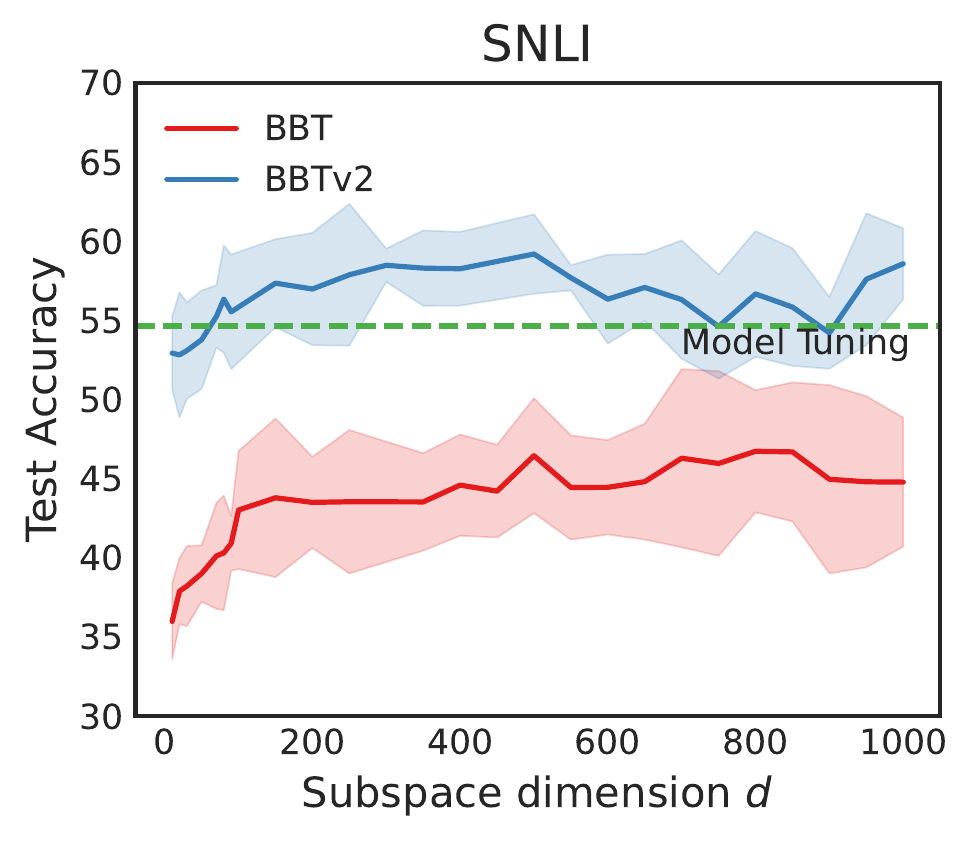}
    \\
    \includegraphics[width=.49\linewidth]{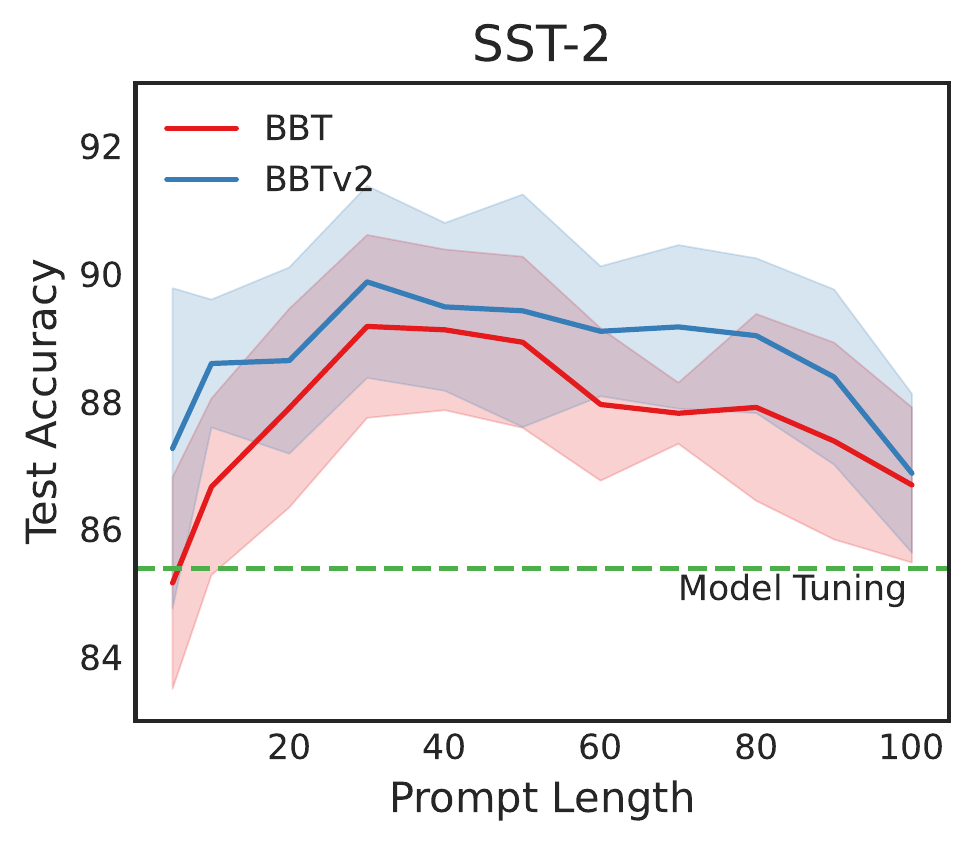}
    \includegraphics[width=.49\linewidth]{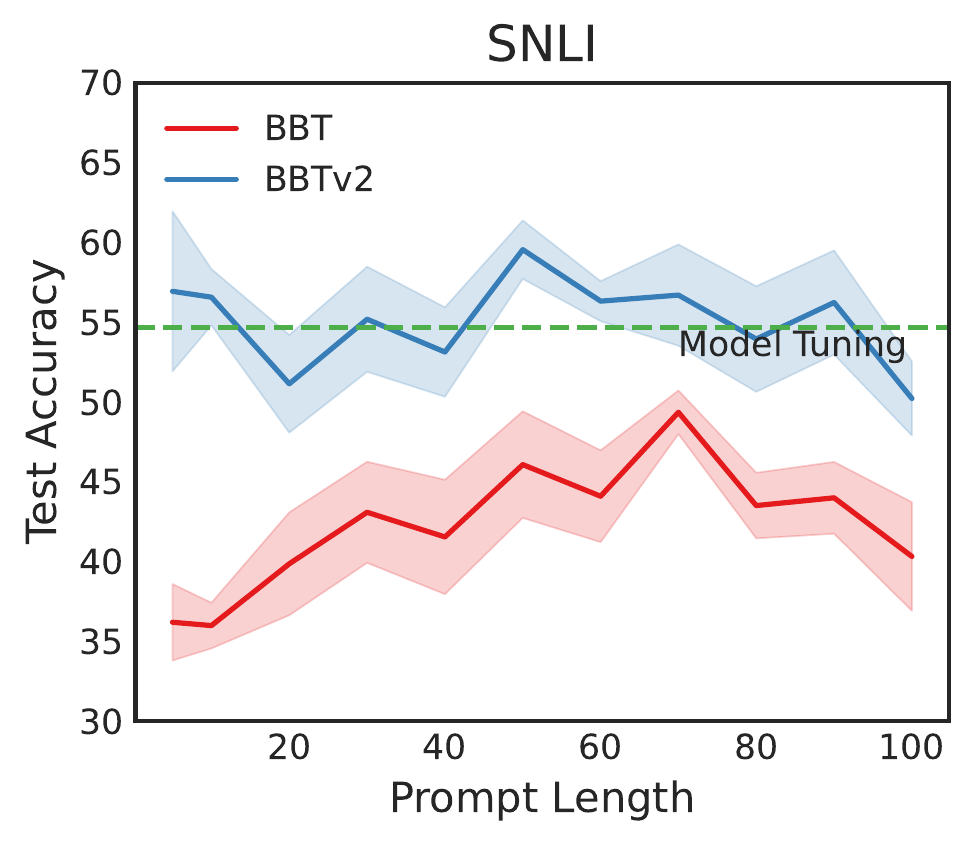}
    \caption{Ablation results on subspace dimensionality and prompt length. We show mean and standard deviation of performance over 5 different runs.}
    \label{fig:ablation}
\end{figure}

\paragraph{Effect of Subspace Dimensionality.}
We explore the subspace dimensionality $d$ from 10 to 1000 using both BBT and BBTv2. The population size is set to $\lambda=4+3\log(d)$ accordingly. Experimental results on SST-2 and SNLI are demonstrated in Figure~\ref{fig:ablation}, from which we observe that: (1) Increasing subspace dimensionality $d$ generally confers improved performance for both BBT and BBTv2, but marginal effect is also observed when $d>100$; (2) BBTv2 almost always performs better than BBT with the same subspace dimensionality.

\paragraph{Effect of Prompt Length.}
As reported in prior work~\cite{Lester2021Prompt,Sun2022BBT}, prompt length can be a sensitive hyperparameter to model performance.
Hence, we explore the prompt length from 5 to 100 using BBT and BBTv2. As shown in Figure~\ref{fig:ablation}: (1) The optimal prompt length lies in the range from 5 to 100 and varies across tasks; (2) The effect of prompt length is somehow consistent between BBT and BBTv2.

\begin{figure}[t]
    \centering
    \includegraphics[width=\linewidth]{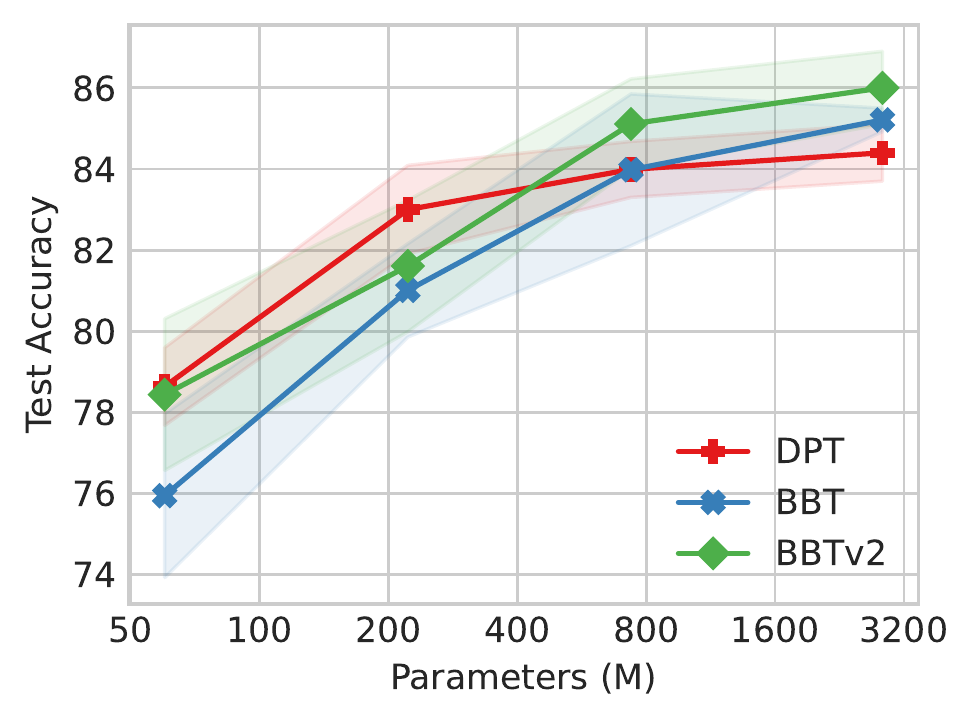}
    \caption{The power of scale for black-box tuning with T5 on AG's News. DPT: deep prompt tuning.}
    \label{fig:t5_agnews}
\end{figure}

\paragraph{Effect of Model Scale.}
It has been demonstrated that larger PTMs have a lower intrinsic dimensionality~\cite{Aghajanyan2021Intrinsic} and therefore, BBT and BBTv2 should be more favorable to larger PTMs. To verify this, we conduct experiments on AG's News using T5~\cite{Raffel2020T5} with varying scales, i.e., T5\textsubscript{SMALL}, T5\textsubscript{BASE}, T5\textsubscript{LARGE}, and T5\textsubscript{XL}, corresponding to 60M, 220M, 740M, and 3B parameters. As shown in Figure~\ref{fig:t5_agnews}, in the AG's News 16-shot learning setting, the gradient-based counterpart, namely deep prompt tuning (DPT, \citet{Li2021Prefix,Liu2021PTuningv2,Qin21Learning}), performs better than BBT and BBTv2 on T5\textsubscript{SMALL} and T5\textsubscript{BASE}. When using T5\textsubscript{LARGE} and T5\textsubscript{XL}, black-box tuning outperforms DPT, demonstrating its power of scale.

\section{Related Work}
\paragraph{Parameter-Efficient Tuning (PET).}
PET is to optimize only a small portion of parameters while keeping the main body of the model unchanged. PET has achieved comparable performance to full model tuning when training data is sufficient~\cite{He2021Unified}. The tunable parameters can be injected into different positions of the PTM. \citet{Houlsby2019Adapter} insert lightweight adapters to each PTM layer; \citet{Lester2021Prompt} prepend continuous prompt tokens to the input layer; \citet{Li2021Prefix,Liu2021PTuningv2} inject tunable prompt tokens to hidden states of every layer; \citet{Zaken22BitFit} only optimize the bias-terms in the PTM; \citet{Hu2021LoRA} learn to adapt attention weights via low-rank matrices. Though the number of tunable parameters is reduced, back-propagation through the entire model is still required to calculate the gradients to update the small portion of parameters. 
To that end, gradient-free methods are also proposed to optimize continuous prompt~\cite{Sun2022BBT,Diao2022Black} or discrete prompt~\cite{Prasad2022GrIPS,Deng2022RLPrompt}.

\paragraph{Prompt-Based Learning.}
Prompt-based learning is to formulate downstream tasks as a (masked) language modeling task, and therefore reduces the gap between PTM pre-training and fine-tuning~\cite{Liu2021PromptSurvey,Sun2021Paradigm}. The prompt can be manually designed~\cite{Brown2020GPT3,Schick20Auto,Schick21PET}, mined from corpora~\cite{Jiang20How}, generated by generative PTMs~\cite{Gao20Making}, or be constructed using gradient-guided search~\cite{Shin20Autoprompt}. In this work, we also insert manually crafted textual prompts into input samples but only optimize the prepended continuous prompt tokens.

\section{Conclusion}
\label{sec:conclusion}
In this work, we present BBTv2, an improved version of BBT~\cite{Sun2022BBT} with deep prompts that are attached to every layer of the PTM. To optimize the high-dimensional prompt parameters, we propose a divide-and-conquer (DC) algorithm combined with random projections to alternately optimize the continuous prompt at each layer. Experimental results demonstrate that BBTv2, without any gradient-based component, can achieve comparable performance to state-of-the-art PET methods and full model tuning while maintaining much fewer tunable parameters.


\section*{Limitations}
We summarize the limitations of this work as follows:
(1) BBTv2 adopts a divide-and-conquer algorithm to alternately optimize prompt at each PTM layer. We use a unique CMA-ES optimizer, which has two hyperparameters $\mu_z$ and $\sigma_z$, for the optimization at each layer. As mentioned previously, we set $\mu_z=0$ for not incorporating any prior to the optimization direction. Therefore, we have totally $L$ (number of layers) hyperparameters for optimization, i.e., $[\sigma_z^{(1)}, \dots, \sigma_z^{(L)}]$. For simplicity, we constrain the $\sigma_z$ at different layers to be identical, i.e., $\sigma_z^{(1)}=\sigma_z^{(2)}=\cdots=\sigma_z^{(L)}$. In addition, Eq.(\ref{eq:sigma}) introduces another hyperparameter $\alpha$, which can also be different across different layers. Similarly, we constrain $\alpha$ at all layers to be identical. Hence, in contrast to BBT that has only one hyperparameter for optimization, BBTv2 introduces an additional hyperparameter and therefore increases the cost for hyperparameter search. (2) We conduct experiments on 9 language understanding tasks across 4 types (i.e., sentiment analysis, topic classification, paraphrasing, and natural language inference) and 2 languages (i.e., English and Chinese). However, the performance of BBTv2 on a wider range of understanding tasks and generation tasks is still under-explored. (3) We limit our work to few-shot learning because the training data can be wrapped into a single batch to be fed into the PTM such that the model inference API is a deterministic function whose output only depends on the prompt. In such a low-noise scenario, we can adopt the CMA-ES to successfully perform optimization. In the full data setting where the training samples are divided into mini-batches, we need to explore other derivative-free optimizers to handle the stochastic (noisy) optimization. We leave the investigation of adapting BBTv2 to a wider range of tasks and to full data settings as future work, further removing the remaining barriers to a gradient-free future.




\section*{Acknowledgements}
This work was supported by National Natural Science Foundation of China (No. 62106076), Natural Science Foundation of Shanghai (No. 21ZR1420300), and National Natural Science Foundation of China (No. 62022027). 

\bibliography{anthology,custom}
\bibliographystyle{acl_natbib}

\newpage
\appendix
\section{Deviation of $\sigma$ for Normal Distribution}
\label{sec:append_A}

Assume the variable $\mathbf{z}\in\mathbb{R}^d$ is sampled from a normal distribution $\mathcal{N}(\mu_z, \sigma_z)$ that is maintained by the CMA-ES, the random projection $\mathbf{A}\in\mathbb{R}^{D\times d}$ is generated from another normal distribution $\mathcal{N}(\mu_A, \sigma_A)$. Considering each entry of the prompt $\mathbf{p}_{ij}=\sum_{k}\mathbf{A}_{ik}\mathbf{z}_{kj}$, the variance is as follows,
\begin{align}
    &\mathbb{V}(\sum_{k=1}^d\mathbf{A}_{ik}\mathbf{z}_{kj}) = \sum_{k=1}^d\mathbb{V}(\mathbf{A}_{ik}\mathbf{z}_{kj}) \nonumber\\
    &\ \ \ \ \ \ \ \  + 2\sum_{k=1}^{d-1}\sum_{p=k+1}^d\mathbb{C}\text{ov}(\mathbf{A}_{ik}\mathbf{z}_{kj}, \mathbf{A}_{ip}\mathbf{z}_{pj}).
\end{align}

Since $\mathbf{A}$ and $\mathbf{z}$, and each entry in $\mathbf{A}$ and $\mathbf{z}$ are independent random variables and therefore the variance of $\sum_{k}\mathbf{A}_{ik}\mathbf{z}_{kj}$ can be simplified as
\begin{align}
    \mathbb{V}(\sum_{k=1}^d\mathbf{A}_{ik}\mathbf{z}_{kj}) &= \sum_{k=1}^d\mathbb{V}(\mathbf{A}_{ik}\mathbf{z}_{kj})\\
    &= d\mathbb{V}(\mathbf{A}_{ik}\mathbf{z}_{kj}). \label{eq:var}
\end{align}
It is easy to obtain that
\begin{align}
    \mathbb{V}(\mathbf{A}_{ik}\mathbf{z}_{kj}) =& \mathbb{V}(\mathbf{A}_{ik})\mathbb{V}(\mathbf{z}_{kj}) +\mathbb{V}(\mathbf{A}_{ik})(\mathbb{E}[\mathbf{z}_{kj}])^2 \nonumber \\
    &+ \mathbb{V}(\mathbf{z}_{kj})(\mathbb{E}[\mathbf{A}_{ik}])^2\\
    =& \sigma_A^2\sigma_z^2 + \sigma_A^2\mu_z^2 + \sigma_z^2\mu_A^2.
\end{align}

Initially, we do not incorporate any prior on the optimization direction of the embedding (or hidden states) and therefore $\mu_A=\mu_z=0$. So we have $\mathbb{V}(\mathbf{A}_{ik}\mathbf{z}_{kj})=\sigma_A^2\sigma_z^2$. Combined with Eq.(\ref{eq:var}), the variance of the entries in the randomly projected prompt is as follows,
\begin{align}
    \mathbb{V}(\mathbf{p}_{ij}) = d\sigma_A^2\sigma_z^2.
\end{align}

Ideally, we expect the generated prompt to lie in a reasonable solution space (e.g., the space of the embedding or hidden states) such that the embedding (hidden states) added by the prompt can still lie in a reasonable space. A natural idea is to match the variance of the generated prompt $\mathbf{p}$ and the embedding (or hidden states). Formally, we expect that $\mathbb{V}(\mathbf{p}_{ij}) = (\alpha\hat{\sigma})^2$, where $\hat{\sigma}$ is the observed standard deviation of the embedding (or hidden states) and $\alpha$ is a constant scalar that controls the range (relative to that of the embedding or hidden states) where $\mathbf{p}$ falls in. We can obtain a recommendation value of the standard deviation of the random projection, that is
\begin{align}
    \sigma_A = \frac{\alpha\hat{\sigma}}{\sqrt{d}\sigma_z}.
\end{align}

In practice, $\alpha$ and $\sigma_z$ are hyperparameters.\footnote{The specific hyperparameters to reproduce the results on each dataset can be found in our code: \url{https://github.com/txsun1997/Black-Box-Tuning}.} It is worth noting that the random projections are static during the optimization process and therefore we only need to observe the standard deviation of the word embeddings and the hidden states at every layer once at the beginning of the optimization. A possible concern is that such observation breaks the black-box and therefore it may be controversial to call it "black-box tuning". We take a perspective of feature-based approaches that views the embeddings and hidden states as features, which are the outputs of the model. Thus, we do not really access the inside information of the black-box model.

\section{Data Preprocessing}
\label{sec:append_data}
\subsection{Statistics of Datasets}
We list the statistics of the 7 English tasks and 2 Chinese tasks used in our experiments in Table~\ref{tab:data}. Among the 9 tasks, 5 are single-sentence classification tasks and 4 are sentence-pair classification tasks. The types of the tasks range from sentiment analysis, topic classification, paraphrase, and natural language inference (NLI).

\begin{table}[h]
\centering
\resizebox{\linewidth}{!}{
\begin{tabular}{llccrrc}
\toprule
\textbf{Category} & \textbf{Dataset} & \textbf{Lang.} & \textbf{$\mid\mathcal{Y}\mid$} & \textbf{$\mid$Train$\mid$} & \textbf{$\mid$Test$\mid$} & \textbf{Type}\\ \midrule
\multirow{5}{*}{\begin{tabular}[c]{@{}l@{}}single-\\ sentence\end{tabular}}
& SST-2 & En & 2 & 67k & 0.9k & sentiment \\
& Yelp P. & En &  2 & 560k & 38k & sentiment \\
& AG's News & En & 4 & 120k & 7.6k & topic \\
& DBPedia & En & 14 & 560k & 70k & topic\\
& ChnSent & Zh & 2 & 6k & 1.2k & sentiment\\ \midrule
\multirow{4}{*}{\begin{tabular}[c]{@{}l@{}}sentence-\\ pair\end{tabular}}
& MRPC & En & 2 & 3.7k & 0.4k & paraphrase \\
& RTE & En & 2 & 2.5k & 0.3k & NLI \\
& SNLI & En & 3 & 549k & 9.8k & NLI \\ 
& LCQMC & Zh & 2 & 239k & 8.8k & NLI \\\bottomrule
\end{tabular}
}
\caption{Statistics of datasets used in our experiments. $\mid\mathcal{Y}\mid$: number of classes. "En" means English and "Zh" means Chinese.}
\label{tab:data}
\end{table}

\subsection{Templates and Label Words}
For both BBT and BBTv2, we convert input texts $X$ with pre-defined templates into $\Tilde{X}$, and convert output labels $Y$ into label words $\Tilde{Y}$, such that downstream tasks can be reformulated into a (masked) language modeling task and therefore we can reuse the pre-trained (masked) language modeling head. In Table~\ref{tab:template}, we list the input and output formats for different PTMs.

\begin{table*}[t]
\centering
\resizebox{\linewidth}{!}{
\begin{tabular}{lll}
\toprule
\textbf{Dataset} & \textbf{Input}             & \textbf{Output}                                                        \\ \midrule
\multicolumn{3}{l}{\textit{Backbone: }RoBERTa, BERT, BART}                                                                                \\ \midrule
SST-2            & $\langle P\rangle$ $\langle S\rangle$. It was \texttt{[MASK]}     & great, bad                                                             \\
Yelp P.          & $\langle P\rangle$ $\langle S\rangle$. It was \texttt{[MASK]}     & great, bad                                                             \\
AG's News         & $\langle P\rangle$ \texttt{[MASK]} News: $\langle S\rangle$       & World, Sports, Business, Tech                                          \\
\multirow{2}{*}{DBPedia} & \multirow{2}{*}{$\langle P\rangle$ [Category: \texttt{[MASK]}] $\langle S\rangle$} & Company, Education, Artist, Athlete, Office, Transportation, Building, \\
                 &                            & Natural, Village, Animal, Plant, Album, Film, Written                  \\
MRPC             & $\langle P\rangle$ $\langle S_1\rangle$ ? \texttt{[MASK]}, $\langle S_2\rangle$    & Yes, No                                                                \\
RTE              & $\langle P\rangle$ $\langle S_1\rangle$ ? \texttt{[MASK]}, $\langle S_2\rangle$    & Yes, No                                                                \\
SNLI             & $\langle P\rangle$ $\langle S_1\rangle$ ? \texttt{[MASK]}, $\langle S_2\rangle$    & Yes, Maybe, No                                                         \\ \midrule
\multicolumn{3}{l}{\textit{Backbone:} GPT-2}                                                                                              \\ \midrule
SST-2            & $\langle P\rangle$ $\langle S\rangle$. The sentiment is  & positive, negative                                                     \\
AG's News        & $\langle P\rangle$ $\langle S\rangle$. The news above is about & world, sports, business, tech                                 \\ 
\multirow{2}{*}{DBPedia} & \multirow{2}{*}{$\langle P\rangle$ $\langle S\rangle$. The text above is about} & company, education, artist, athlete, office, transportation, building, \\
& & natural, village, animal, plant, album, film, written \\ \midrule
\multicolumn{3}{l}{\textit{Backbone:} T5}                                                                                              \\ \midrule
SST-2            & $\langle P\rangle$ $\langle S\rangle$. It was \texttt{[X]}     & \texttt{[X]} positive/negative                                                             \\
AG's News         & $\langle P\rangle$ \texttt{[X]} News: $\langle S\rangle$       & \texttt{[X]} World/Sports/Business/Tech                                          \\
DBPedia          & $\langle P\rangle$ [Category: \texttt{[X]}] $\langle S\rangle$ & \texttt{[X]} Company/Education/Artist/Athlete/Office/Transportation/Building/... \\ \midrule
\multicolumn{3}{l}{\textit{Backbone:} CPM-2}                                                                                              \\ \midrule
ChnSent & \begin{CJK}{UTF8}{gbsn}$\langle P\rangle$ $\langle S\rangle$。总之很\texttt{[X]}。\end{CJK} & \begin{CJK}{UTF8}{gbsn}\texttt{[X]} 好 / 差\end{CJK}\\
LCQMC & \begin{CJK}{UTF8}{gbsn}$\langle P\rangle$ 判断：$\langle S_1\rangle$和$\langle S_2\rangle$两句话的意思是\texttt{[X]}的。\end{CJK} & \begin{CJK}{UTF8}{gbsn}\texttt{[X]} 矛盾 / 相似\end{CJK}\\
\bottomrule
\end{tabular}
}
\caption{Input templates and output label words used in our experiments. $\langle P\rangle$ is a sequence of continuous prompt tokens. $\langle S\rangle$ is the original input text. For BART, which outputs denoised input in an auto-regressive fashion, we only use the prediction of the masked position such that it follows the same output format as BERT and RoBERTa. For T5 and CPM-2, \texttt{[X]} is a special token similar to \texttt{[MASK]}.}
\label{tab:template}
\end{table*}

\section{Implementation Details}
\label{sec:append_implement}


\paragraph{Clipping Hidden States}
In practice, we find that the standard deviation $\hat{\sigma}$ of hidden states (especially at high layers) in some PTMs can be very large due to a few outliers. As a result, our calculated $\sigma_A$ (Eq.(\ref{eq:sigma})) becomes large accordingly, and therefore the generated prompt has a wider range of values than expected. To address this issue, we iteratively clip hidden states into the range of $\hat{\mu}\pm 3\hat{\sigma}$. We perform clipping for 5 rounds and then compute the standard deviation $\hat{\sigma}$ of hidden states. Note that the clipping is only performed for statistics but is not applied during model forward compute.

\section{Additional Results}
\label{sec:append_add_res}

\begin{figure}[t]
    \centering
    \includegraphics[width=.49\linewidth]{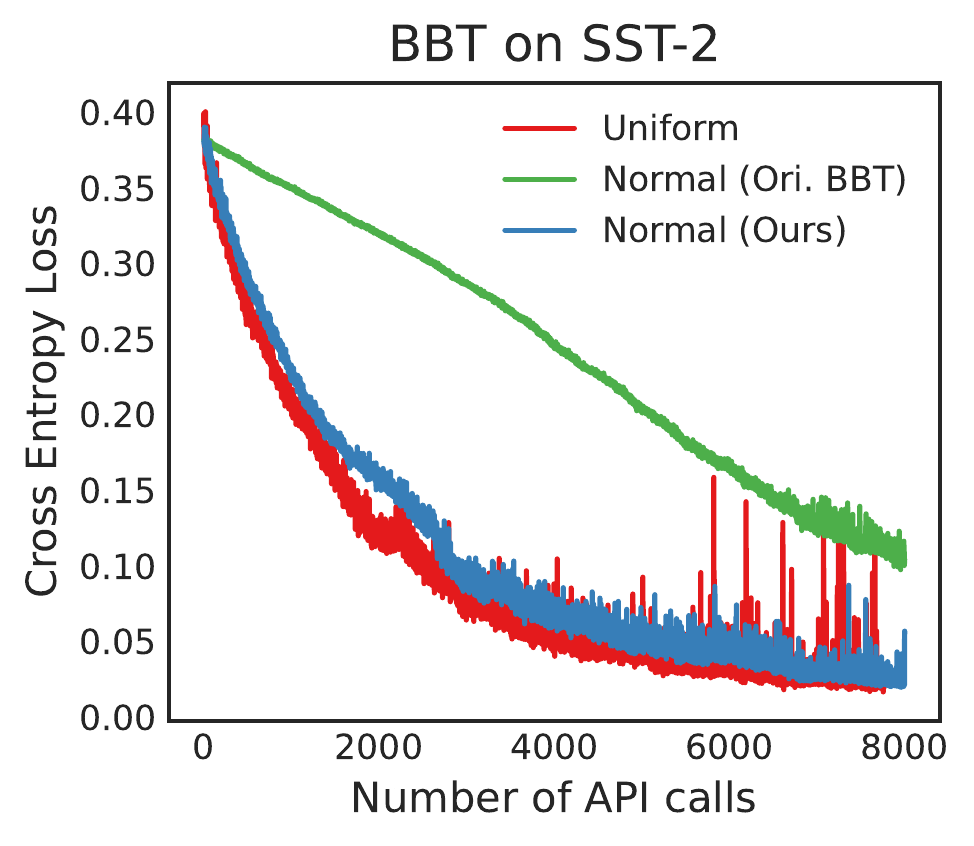}
    \includegraphics[width=.49\linewidth]{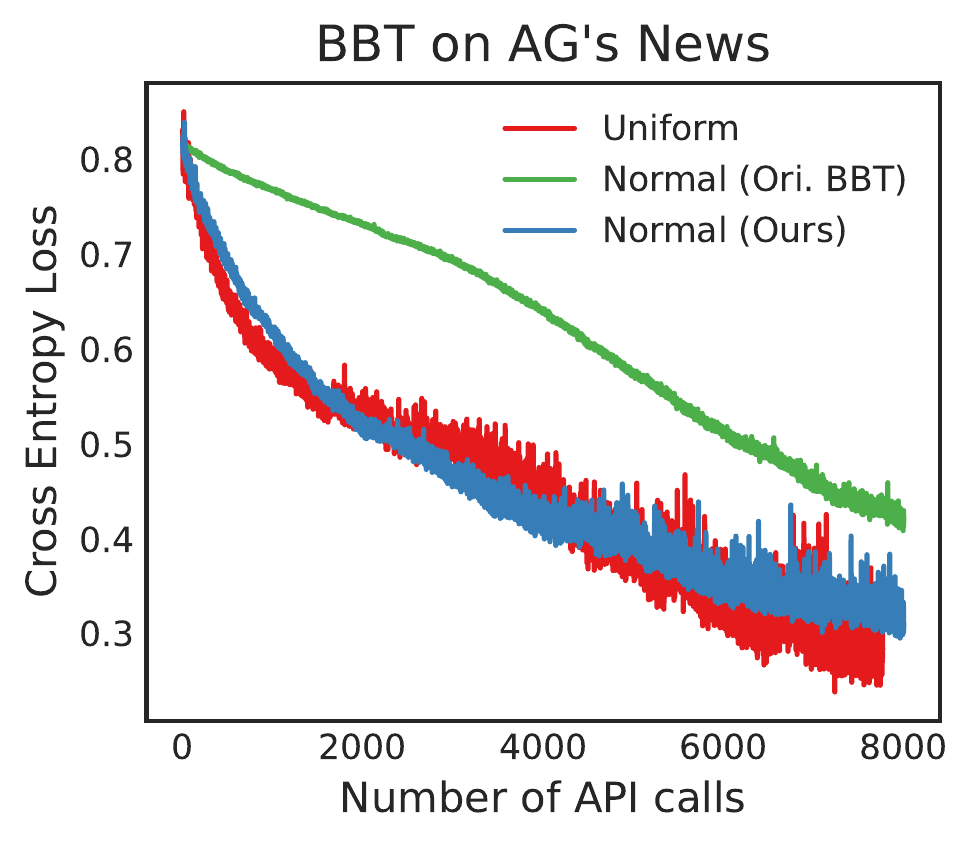}
    \\
    \includegraphics[width=.49\linewidth]{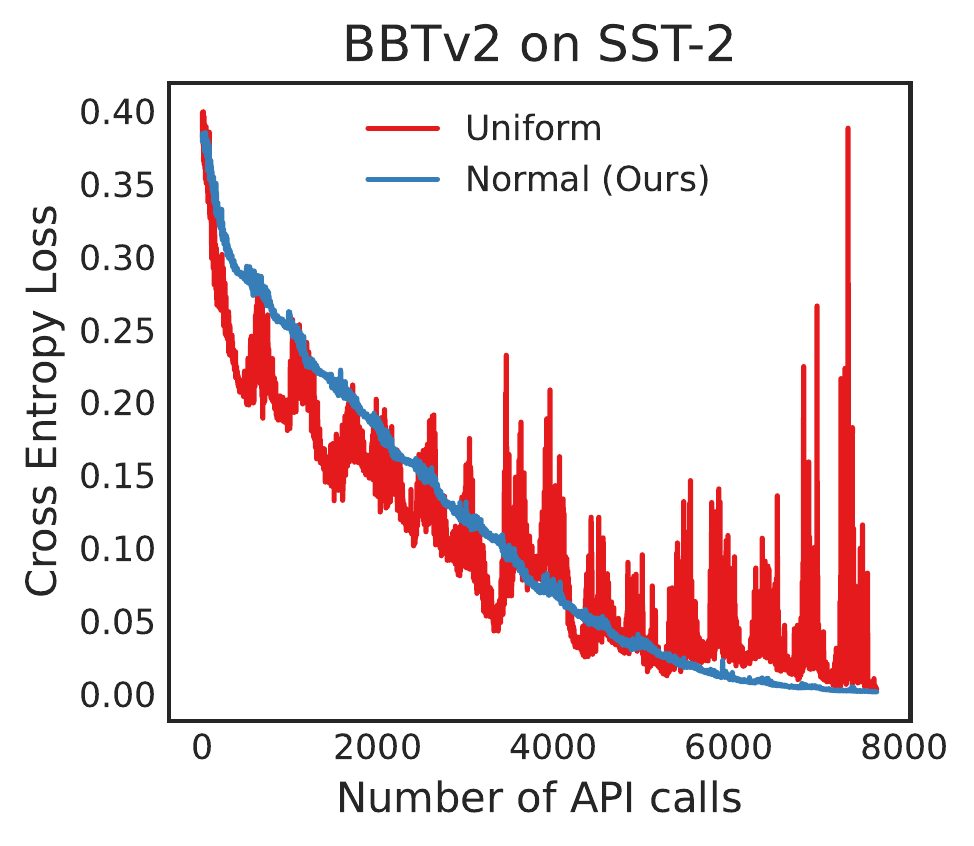}
    \includegraphics[width=.49\linewidth]{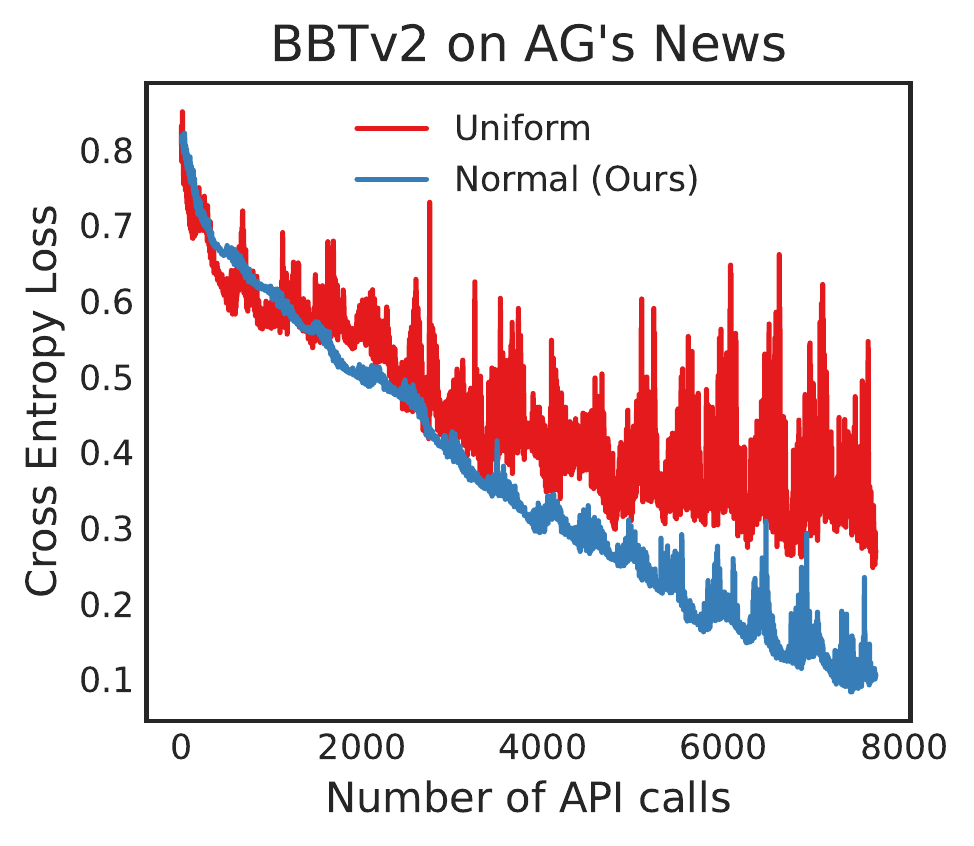}
    \caption{Comparison of convergence rates with random projections with different distributions. When using our designed normal distribution to generate random projections, both BBT and BBTv2 achieve fast and stable convergence.}
    \label{fig:re}
\end{figure}

\paragraph{On Convergence of Normal Distributions}
Previously in Figure~\ref{fig:overfit}, we show that using our designed normal distribution leads to better generalization from training data to development data. Nevertheless, as reported by \citet{Sun2022BBT}, using normal distributions can suffer from slow convergence. Therefore, we compare the convergence rates using random projections generated from different distributions. As demonstrated in Figure~\ref{fig:re}: (1) For BBT, the convergence rate of using our designed normal distribution is significantly faster than the normal distribution used in the original BBT, and is comparable to uniform distribution; (2) For BBTv2, using our normal distribution converges more stably on both SST-2 and AG's News. Especially, we observe that using our normal distribution converges faster than uniform distribution on AG's News.

\section{Estimation of Memory Use and Network Load}
\label{sec:append_network_load}
For measurement of memory footprint on user side, we use \texttt{psutil} to monitor CPU memory when running CMA-ES. For memory footprint on service side, we use \texttt{nvidia-smi} to monitor GPU memory when serving PTM inference.

For estimation of network load, we measure the amount of data to be uploaded and downloaded. For BBT and BBTv2, there are two kinds of data to be uploaded: (1) training samples, and (2) continuous prompt. A training sample is comprised of two parts: \texttt{input\_ids} and \texttt{attention\_mask}. We can use the unsigned short (representation range: 0$\sim$65535, 2 bytes per value) for \texttt{input\_ids} and use the bool type (1 byte per value) for \texttt{attention\_mask}. For continuous prompt, which contains hundreds of values for BBT or tens of thousands of values for BBTv2, we can use the float type (4 bytes per value) for representation. Take SST-2 16-shot split as an example, the \texttt{input\_ids} and \texttt{attention\_mask} are in shape of $32\times 47$, where 32 is the batch size and 47 is the maximum sequence length, so there are $\sim$2.9KB data for \texttt{input\_ids} and $\sim$1.5KB data for \texttt{attention\_mask}. Assume the subspace dimensionality is 500, we need to upload additional $\sim$2KB data for prompt if using BBT and $\sim$48KB data if using BBTv2. The data to be downloaded is the output logits of the candidate words, which is a dictionary containing $\mid\mathcal{Y}\mid$ float values. Take SST-2 16-shot split as an example, the size of data to be downloaded is $32\times2\times4\text{bytes}=0.25$KB. We assume the random projections are generated on the server side therefore there is no need to download hidden states to compute standard deviations for users.
\end{document}